\documentclass[10pt,twocolumn,letterpaper]{article}

\usepackage{iccv}              

\usepackage{microtype}
\usepackage[sort&compress,numbers]{natbib}
\usepackage{mathrsfs}
\usepackage{amsmath}
\usepackage{bm}
\usepackage{makecell}
\usepackage{multirow}
\usepackage{adjustbox}
\usepackage{graphicx}
\usepackage{colortbl}
\usepackage{arydshln}
\usepackage{microtype} 
\usepackage{amsmath, amssymb, mathtools}
\usepackage{appendix}
\usepackage{wrapfig}
\usepackage[ruled,vlined]{algorithm2e}
\usepackage{amsfonts} 
\usepackage{subcaption} 
\definecolor{mygray}{gray}{0.6}
\definecolor{cvprblue}{rgb}{0.21,0.49,0.74}
\definecolor{iccvblue}{rgb}{0.21,0.49,0.74}
\usepackage[pagebackref,breaklinks,colorlinks,allcolors=iccvblue]{hyperref}
\usepackage[capitalize]{cleveref}

\newcommand\nnfootnote[1]{%
  \begin{NoHyper}
  \renewcommand\thefootnote{}\footnote{#1}%
  \addtocounter{footnote}{-1}%
  \end{NoHyper}
}

\def\paperID{5411} 
\def\confName{ICCV}
\def\confYear{2025}

\title{CalibQuant: 1-Bit KV Cache Quantization for Multimodal LLMs}

\author{~\textsuperscript{$\star$}Insu Han$^{1}$, ~\textsuperscript{$\star$}Zeliang Zhang$^{2}$, ~\textsuperscript{$\star$}Zhiyuan Wang$^{3}$, ~\textsuperscript{$\star$}Yifan Zhu$^{2}$,\\
{Susan Liang}$^{2}$, {Jiani Liu}$^{2}$, {Haiting Lin}$^{4}$, {Mingjie Zhao}$^{5}$, Chenliang Xu$^{2}$, \textsuperscript{$\dagger$}Kun Wan$^{4}$, \textsuperscript{$\dagger$}Wentian Zhao$^{4}$ \\
  $^1$KAIST \quad \quad $^2$University of Rochester \quad \quad $^3$UCSB \quad \quad $^4$Adobe Inc. \quad \quad $^5$ Independent Researcher\\
  \texttt{insu.han@kaist.ac.kr}, \\
  \texttt{ \{zeliang.zhang, yifan.zhu, susan.liang, chenliang.xu\}@rochester.edu},  \\
  \texttt{jliu186@u.rochester.edu}, 
  \texttt{\{wezhao, kuwan, halin\}@adobe.com}\\
  \texttt{zwang796@ucsb.edu}, \texttt{mjzhao1@gmail.com}
}

\begin{document}
\maketitle

\begin{abstract}
\nnfootnote{$\star$ indicates the equal contribution with random order.}
\nnfootnote{$\dagger$ indicates the project leaders.}
Multimodal Large Language Models (MLLMs) have demonstrated remarkable performance across diverse applications. However, their computational overhead during deployment remains a critical bottleneck. While Key-Value (KV) caching effectively trades memory for computation to enhance inference efficiency, the growing memory footprint from extensive KV caches significantly reduces throughput and restricts prolonged deployment on memory-constrained GPU devices. To address this challenge, we propose CalibQuant, a simple yet highly effective visual quantization strategy that drastically reduces both memory and computational overhead. Specifically, CalibQuant introduces an extreme 1-bit quantization scheme, complemented by novel post-scaling and calibration techniques tailored to the intrinsic patterns of KV caches, thereby ensuring high efficiency without compromising model performance. Leveraging Triton for runtime optimization, we achieve a {\bf 10x} throughput increase on InternVL models. Our method is designed to be plug-and-play, seamlessly integrating with various existing MLLMs without requiring architectural changes. Extensive experiments confirm that our approach significantly reduces memory usage while maintaining computational efficiency and preserving multimodal capabilities. Codes are available at \href{https://github.com/insuhan/calibquant}{\url{https://github.com/insuhan/calibquant}}.
\end{abstract}

\section{Introduction}

Multimodal Large Language Models (MLLMs) have demonstrated strong performance across a wide range of tasks including automated caption generation, interactive storytelling, medical image diagnostics and emotion-driven captioning, to name a few~\citep{tang2023video,bai2023qwen,chen2024mllm}. 
However, due to the quadratic computation complexity and linear memory complexity of the self-attention mechanism~\cite{bahdanau2014neural}, Transformer-based MLLMs present significant challenges in terms of memory consumption as the number of visual frames and the image resolution increase~\citep{zhang2024treat}. The resulting surge in visual tokens further amplifies the computational burden, making the deployment of MLLMs in real-world applications increasingly difficult. To address these challenges and accelerate MLLMs, various approaches have been proposed to reduce computational costs and improve throughput. These include developing compact multimodal language models~\citep{abdin2024phi}, applying model pruning~\citep{zhang2024treat,zhang2024diversifying}, leveraging mixture-of-experts strategies~\citep{dai2024deepseekmoe,jiang2024mixtral}, and optimizing KV cache mechanisms~\citep{wan2024look,zhang2023h2o,zandieh2024qjl,zandieh2024subgen,han2025polarquant,han2025balancekv}.

Among these acceleration methods, KV cache optimization has gained widespread popularity due to its scalability across different models. 
By storing and reusing intermediate key and value (KV) states during decoding, it allows us to avoid operations running in quadratic time in the number of tokens.
The KV cache trades memory for computational efficiency.
However, as the size of the KV cache increases linearly in the number of generated tokens, it causes a memory bottleneck.


In this work, we study an approach to quantize the KV cache in MLLMs, i.e., retaining all token embeddings while storing them in a low-bit format.
We begin with the well-known uniform integer quantization technique~\cite{wu2020integer}, widely adopted for its simplicity. However, when applied to MLLMs, global uniform quantization across the KV cache fails to capture the distinct distributional properties of visual tokens, leading to significant quantization errors, especially at extremely low-bit levels. To address this, we apply channel-wise quantization for the key cache. This approach has been previously observed in \cite{liukivi} for LLMs, where a small subset of channels in key cache contains outliers, making channel-wise quantization particularly effective. However, to the best of our knowledge, this is the first observation of its superiority in MLLMs, particularly for handling visual tokens, where distributional variations are more pronounced. Additionally, we propose a post-scaling trick that leverages the linearity of the dequantization process to restore the key cache from low-bit precision to full precision. We defer these operations but apply a similar transformation to the query state. Since the query represents only a single token during decoding, this approach significantly improves computational efficiency. 

We further observe that outliers in the KV cache can distort the attention mechanism, as extreme values are overrepresented post-quantization, degrading performance. To mitigate this, we introduce a novel post-quantization calibration strategy that adjusts pre-softmax attention scores by aligning their distributions with unquantized baselines, effectively reducing the impact of extreme value distortions. Our experiments show that after applying this calibration, the distribution of pre-softmax attention scores closely matches the unquantized baseline, yielding performance nearly identical to full-precision models. These contributions collectively enable efficient, low-bit KV cache quantization tailored for MLLMs, preserving multimodal reasoning capabilities while significantly reducing memory overhead.


Our contributions can be summarized as follows:

\begin{itemize}
    \item We introduce a novel 1-bit quantization for the visual KV cache in MLLMs, demonstrating superior performance across diverse tasks such as image captioning (COCO Caption), video understanding (MMBench-Video), and document visual question answering (DocVQA). 
    \item Our method builds on uniform integer quantization applying channel-wise scheme for both key and value caches. Additionally, we propose a post-quantization calibration technique to adjust pre-softmax attention scores, effectively reducing the impact of extreme values and improving not only approximation quality but also the end-to-end performance.
    \item We implement our quantization algorithm using the Triton kernel, achieving significant acceleration in the decoding stage with a speedup of up to 11.24$\times$ compared to the 16-bit baseline. In particular, we introduce a post-scaling trick further optimizes computational efficiency by deferring dequantization, reducing memory overhead. 
\end{itemize}

\section{Related Work}

\paragraph{Efficient Inference of MLLMs.}
Multimodal large language models (MLLMs) typically contain billions of parameters, posing significant challenges in both memory consumption and computational efficiency during deployment. Numerous studies have explored cost reduction strategies for MLLM deployment, including designing compact multimodal models~\citep{lin2024moe,abdin2024phi,chu2023mobilevlm,chu2024mobilevlm,li2024mini, yao2024minicpm}, model pruning~\citep{zhang2024treat,shang2024llava-prumerge,xing2024pyramiddrop,chen2024fastv}, and hardware-software co-optimization~\citep{kwon2023efficient}. However, the self-attention mechanism, which has quadratic computational complexity, remains a bottleneck~\citep{vaswani2017attention}. As input sequence length increases, both memory usage and computational burden grow correspondingly. During decoding, every generated token involves computations over all preceding input tokens, exacerbating inefficiencies.

\paragraph{KV Cache Compression.}
The KV cache technique has been introduced to mitigate redundant computations~\citep{zhang2023h2o}. By caching key and value embeddings in memory, the KV cache allows the model to reuse stored information instead of recomputing attention scores for all previous tokens. This approach effectively trades off memory usage for computational efficiency, significantly improving inference speed. 
While the KV cache technique substantially reduces computational overhead, it introduces a new bottleneck: memory consumption. This issue becomes increasingly critical in scenarios involving long-context generation and multi-turn conversations, where growing input lengths negatively impact throughput.

A common approach to reducing the size of the KV cache is to remove or evict unimportant key and value vectors from the cache.
A line of research explores various importance scores to evict tokens~\cite{zhang2023h2o, jin2024llm, cai2024pyramidkv}. 
Quantization techniques provide another path to reducing KV cache memory overhead by transforming floating-point representations into lower-precision integers, thus significantly reducing memory usage and potentially enhancing inference speed. Methods such as KIVI~\citep{liukivi} have introduced asymmetric quantization strategies.
KVQuant~\citep{hooper2024kvquant} introduces advanced KV cache quantization techniques, including per-channel, pre-RoPE, non-uniform, and per-vector quantization, significantly improving accuracy at low bitwidths.
QJL~\citep{zandieh2024qjl} leverages a Johnson-Lindenstrauss (JL) transform combined with sign-bit quantization to eliminate the storage overhead associated with quantization constants in KV cache quantization.
In contrast to these approaches developed primarily for general-purpose LLMs, our method specifically addresses the unique challenges of MLLMs, where visual tokens dominate the KV cache and present distinct statistical patterns. By carefully exploiting the distributional characteristics of visual activations, we introduce a specialized channel-wise quantization combined with attention-aware calibration, significantly reducing memory footprint while preserving multimodal reasoning capabilities.

\section{Preliminaries}


\subsection{KV Cache for Efficient Token Generation}
In autoregressive sequence generation with Transformers~\citep{vaswani2017attention}, the key-value (KV) cache improves efficiency by eliminating redundant computation in self-attention. The token generation process consists of two stages: \textbf{prefill} and \textbf{decoding}.  

In the \textbf{prefill} stage, the model processes a prompt of length $n$ and computes key and value for all tokens in a form of matrices 
$K \in \mathbb{R}^{n \times d}, V \in \mathbb{R}^{n \times d}$
where $d$ is the embedding dimension. These caches store past key and value embeddings, enabling efficient self-attention without recomputation in the next token generation steps.

In the \textbf{decoding} stage, the model generates one token at a time. Given a query vector $q_{\text{new}} \in \mathbb{R}^{1 \times d}$ and its corresponding key-value pair $(k_{\text{new}}, v_{\text{new}})$, the KV cache updates by appending the new key and value:  
\begin{align}
K \leftarrow [K; k_{\text{new}}], \quad V \leftarrow [V; v_{\text{new}}].
\end{align}  
The attention output is then computed as:  
\begin{align}
\text{softmax} \left( \frac{q_{\text{new}} K^T}{\sqrt{d}} \right) V.
\end{align}  
This incremental update reduces the time complexity of self-attention from \( O(n^2 d) \) to \( O(n d) \) per step. While this explanation considers a single layer and head, the same mechanism applies across multiple layers and heads in practical Transformer models.

\subsection{Uniform Integer Quantization} \label{sec-uniform-quant}

Quantization is a process that reduces high-precision floating-point values (e.g., 32-bit floating point) to lower-precision integer representations (e.g., 8-bit integer). This compression not only reduces memory space but also accelerates computation speed, which is particularly useful in resource-constrained environments like edge devices and accelerators.  

A widely used approach is the uniform integer quantization. Given a bitwidth $b>0$ and an input value $x$ within a range $[ \alpha, \beta ]$ for some $\beta > \alpha$, it is mapped to a discrete integer $x_{\text{dis}} \in \{0, 1, \dots, 2^b-1\}$ computed as
\begin{align}
    x_{\text{dis}} = \left\lfloor\left(x  - \alpha\right)\cdot \frac{2^b - 1}{\beta - \alpha}\right\rceil, \label{eq-discretize}
\end{align}  
where \( \lfloor \cdot \rceil \) denotes the rounding operator. Representing $x_{\text{dis}}$ requires $b$ bits and it can reduce memory requirements significantly when $b$ is smaller than the full precision.

To recover an approximate floating-point representation, the dequantization process converts it back as follows:  
\begin{align}
    x_{\text{deq}} = x_{\text{dis}} \cdot \frac{\beta - \alpha}{2^b - 1} + \alpha.\label{eq-recover}
\end{align}  
This can be naturally extended to vectors or matrices by applying entry-wise.

\section{CalibQuant: Low-Bit Quantization of Visual KV Cache via Calibration}

Tokens in multimodal large language models (MLLMs) encompass multiple modalities.  For instance, vision-language models like InternVL process both textual and visual tokens. In this work, we focus on scenarios where visual tokens dominate, meaning their sequence length exceeds that of textual tokens such as image captioning and video understanding tasks involving text. In these cases, the KV cache for visual tokens becomes a memory bottleneck. To improve the efficiency of token generation, we propose applying the uniform integer quantization to the visual KV cache.

Given a KV cache, we first determine appropriate values $\alpha$ and $\beta$ that serve as the lower and upper bounds for all entries in the cache. The cache is then encoded into $b$-bit representations, where the bitwidth $b$ controls the trade-off between memory efficiency and attention accuracy; smaller $b$ values lead to greater information loss. Our primary objective is to quantize visual KV caches to low-bit precision, such as $b=2$ or $1$, while minimizing performance degradation. To improve the accuracy of low-bit quantization in the visual cache, we incorporate two simple yet effective strategies: channel-wise quantization and calibration.

\subsection{Channel-wise Quantization with Post-scaling} 

In order to apply the quantization in \cref{eq-discretize}, one needs to determine the minimum and maximum values of the target vectors, i.e., $\alpha$ and $\beta$. While a na\"ve approach would compute these extreme values using global statistics, we instead refine the statistical range along the channel axis.
Specifically, let $K \in \mathbb{R}^{n \times d}$ be a key cache where $n$ and $d$ denote the number of tokens and the head dimension, respectively. We define vectors $\alpha, \beta \in \mathbb{R}^d$ as:
\begin{align}
    \alpha_i = \min_{j \in [n]} K_{j,i}, \quad \beta_i = \max_{j \in [n]} K_{j,i}
\end{align}
for each $i \in [d]$. Each row vector in $K$ is then quantized using \cref{eq-discretize} and \eqref{eq-recover}, where the multiplication is applied entry-wise. We similarly implement this channel-wise uniform quantization approach for the value cache. 

Channel-wise quantization was previously studied by \citet{liukivi} and introduced as KIVI. However, their approach applied channel-wise quantization to the key cache and a token-wise method to the value cache in language models. In contrast, for MLLMs, we find that applying channel-wise quantization to both the key and value caches yields superior performance compared to KIVI. Additionally, we verify that global quantization of the value cache degrades performance compared to our approach. See \cref{sec-experiments} for details.

\paragraph{Post-scale for Efficiency Key Cache Management.}
The quantized key cache can be represented by discretized integer values, a scale factor and a bias term. 
During the decoding stage, these components are utilized to dequantize and reconstruct the key cache, which is subsequently multiplied by the query. However, channel-wise quantization requires distinct scale and bias vectors, resulting in numerous unique values and increased computational overhead during dequantization. 
In addition, this makes computations in the CUDA kernel inefficient. By observing that the quantized keys have a limited set of discrete values (e.g., 0,1,2,3 for 2-bit quantization), we leverage a simple algebraic rearrangement to reduce storage and improve computational efficiency.

More formally, let $k\in\mathbb{R}^d$ be any row vector within the key cache $K \in \mathbb{R}^{n \times d}$ and $k_{\text{dis}}$ be its $b$-bit integer quantization, accompanied by channel-wise scaling factors $\alpha, \beta\in\mathbb{R}^d$. Given a query $q\in\mathbb{R}^d$, the attention in token generation requires the computation of:
\begin{align}
    q \cdot k_{\text{deq}} &= q \cdot \left( k_{\text{dis}} \odot \frac{\beta-\alpha}{2^b-1} + \alpha\right) \nonumber \\
    &= \left(q \odot \frac{\beta-\alpha}{2^b-1}\right) \cdot k_{\text{dis}} + q\cdot \alpha
\end{align}
where $\cdot$ and $\odot$ denote the inner-product and entry-wise product between vectors, respectively. In particular, the channel-wise dequantization operation $\left( k_{\text{dis}} \odot \frac{\beta-\alpha}{2^b-1} + \alpha\right)$ is deferred and efficiently integrated into the subsequent vector multiplications. Furthermore, this approach stores only the $b$-bit integer quantized values, avoiding full-precision dequantization computations whose dimension scales with token length $n$, thereby substantially reducing memory requirements. Consequently, this approach ensures the efficiency of low-bit uniform quantization with channel-wise scaling. The post-scale approach can be naturally applied to the dequantization of the value cache.

\begin{figure*}[htbp]
    \centering
    \begin{subfigure}{0.24\textwidth}
        \centering
        \includegraphics[width=\linewidth]{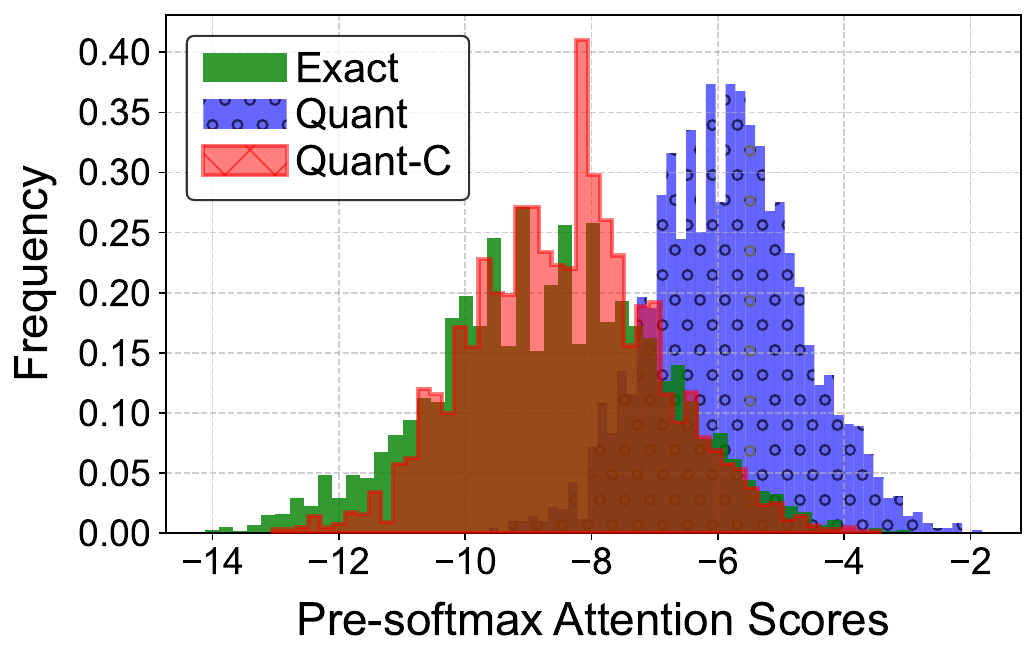}
    \end{subfigure}
    \begin{subfigure}{0.24\textwidth}
        \centering
        \includegraphics[width=\linewidth]{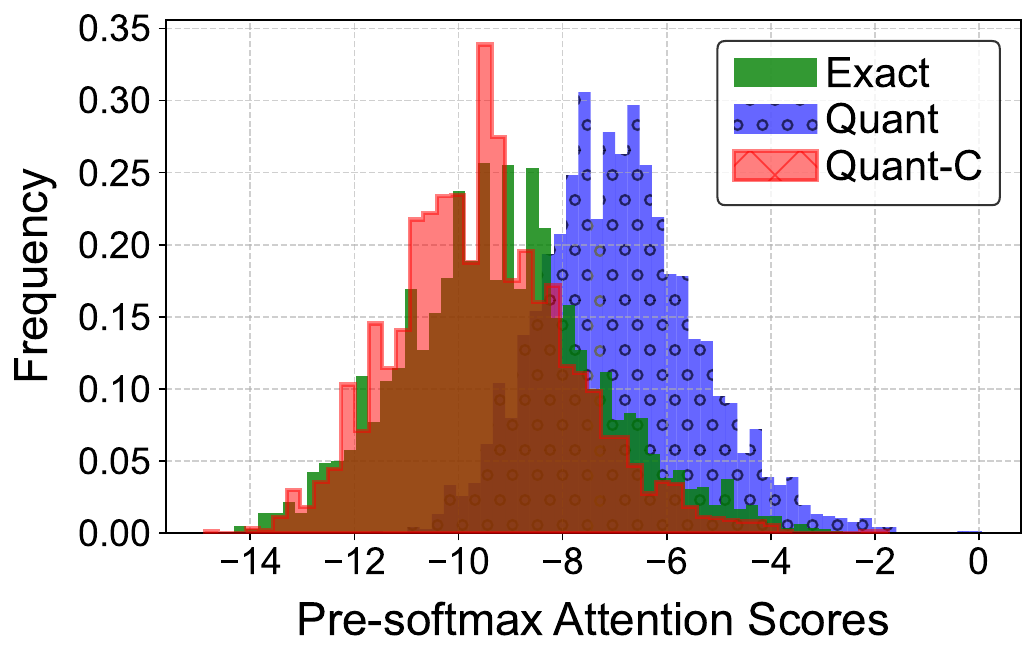}
    \end{subfigure}
    \begin{subfigure}{0.24\textwidth}
        \centering
        \includegraphics[width=\linewidth]{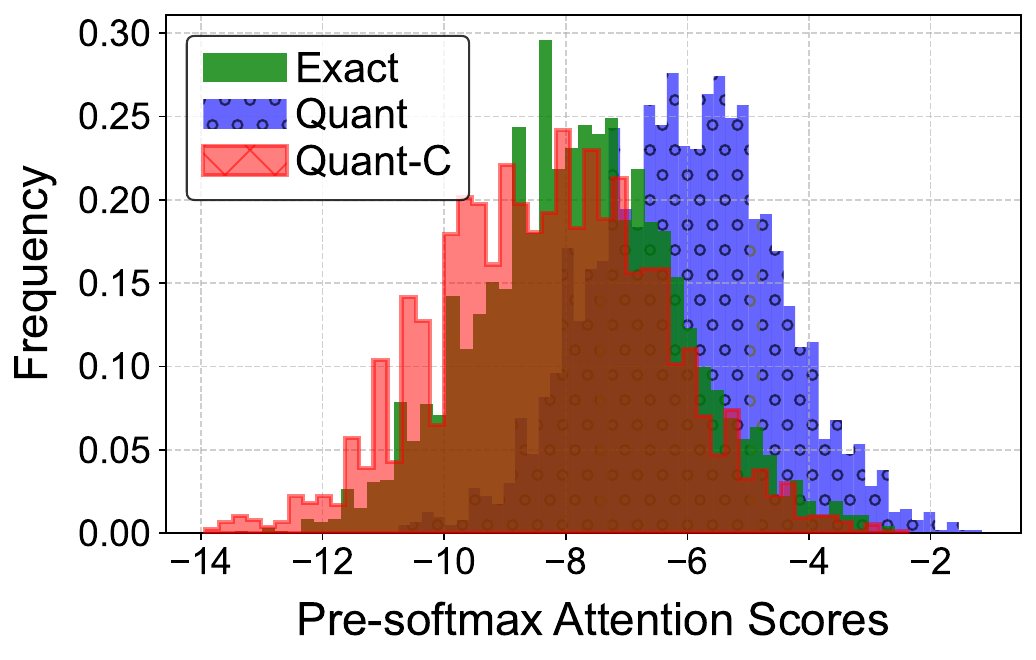}
    \end{subfigure}
    \begin{subfigure}{0.24\textwidth}
        \centering
        \includegraphics[width=\linewidth]{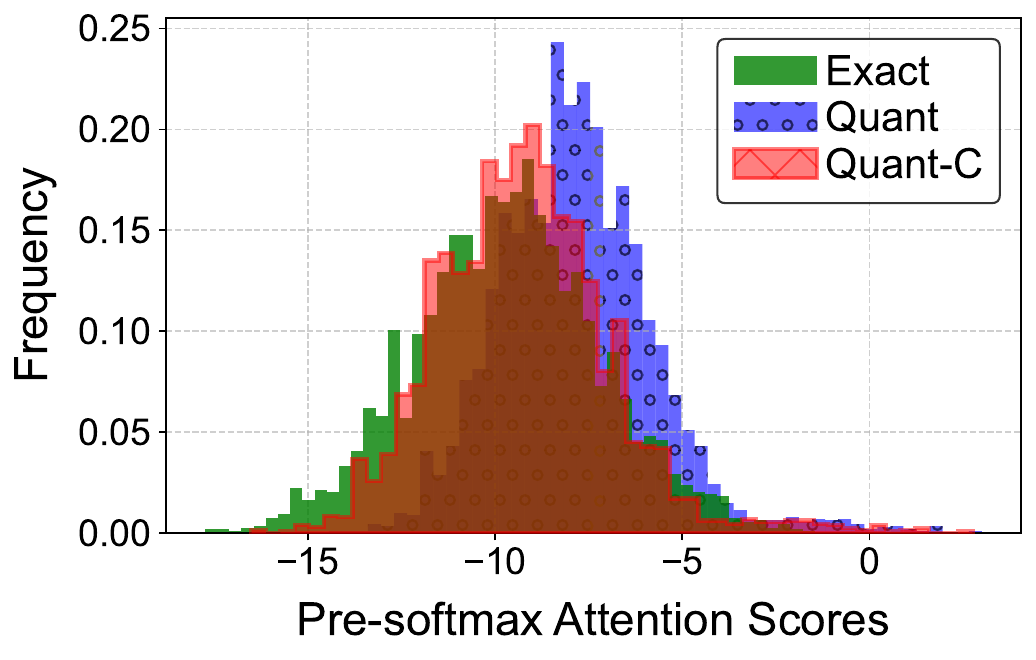}
    \end{subfigure}
    \caption{Distribution of entries in $q K^T/\sqrt{d}$ without quantization (Exact, green), with quantization (Quant, blue) and calibration on post-quantization (Quant-C, red) across different layers and heads.}
    \label{fig-qk-dist}
\end{figure*}

\begin{figure}
    \centering
    \includegraphics[width=0.7\linewidth]{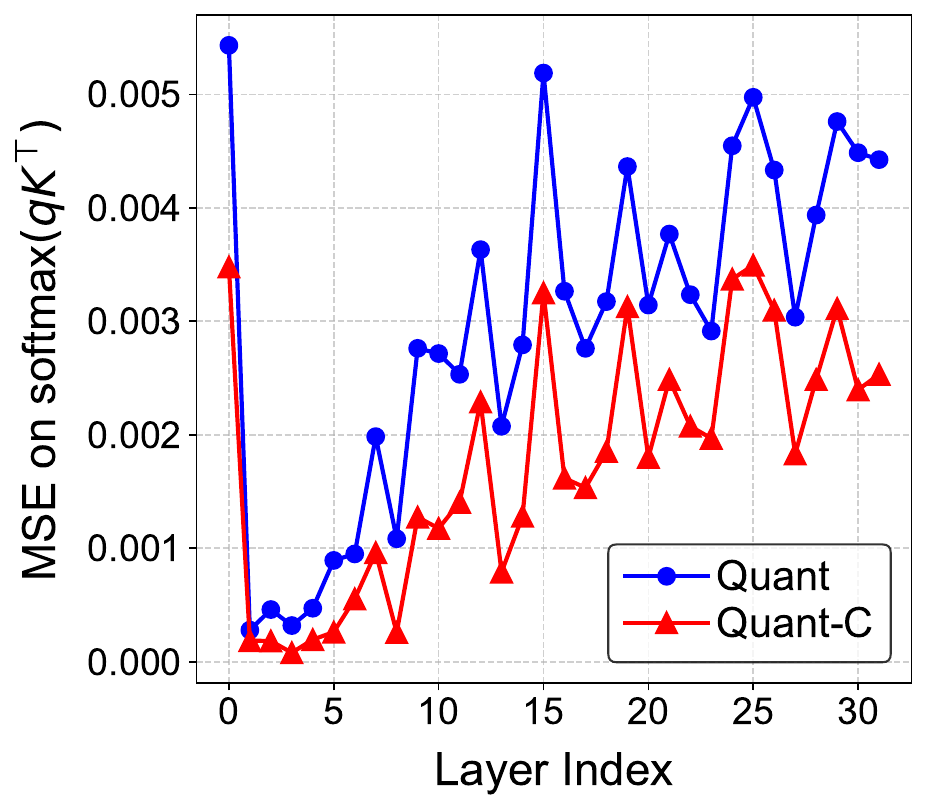}
    \vspace{-0.05in}
    \caption{Mean squared error (MSE) for $\mathrm{softmax}(q K^\top/\sqrt{d})$ across multiple layers. The quantization with calibration (Quant-C, red) shows much lower errors than the quantization only method (Quant, blue).}
    \label{fig-approx-error}
\end{figure}

\subsection{Calibration of Post-quantization} \label{sec-calibration}
A crucial drawback of uniform quantization, as discussed in \cref{sec-uniform-quant}, is that the dequantized values tend to contain a larger number of extreme values. 
This arises because the quantization codebook always includes the minimum and maximum values, causing inputs that are closest to these bounds to be mapped to these extremes upon dequantization. For example, when $b=1$ (1-bit quantization), each element in the dequantized vector $x_{\text{deq}}$ is restricted to be either $\alpha$ or $\beta$.
Consequently, the reconstructed KV cache often contains disproportionately large absolute values, resulting in distorting the output of attentions. 

To address this issue, we propose a novel post-quantization calibration that adjusts the peak values of pre-softmax attention scores.
More precisely, consider a query vector $q \in \mathbb{R}^{d}$ and a key cache $K \in \mathbb{R}^{n \times d}$, with $K_{\text{deq}}$ denoting the dequantized key cache.
Let $K_{\text{deq}}$ be the dequantization of $K$. 
The pre-softmax attention scores are computed as $q K_{\text{deq}}^T/\sqrt{d}$.
We investigate empirical distributions of the elements in both $q K^T/\sqrt{d}$ and $q K_{\text{deq}}^T/\sqrt{d}$, using InternVL2.5-8B model on COCO Caption dataset~\cite{chen2015microsoft}, applying 1-bit quantization to the visual key cache.
\cref{fig-qk-dist} illustrates normalized histograms of the pre-softmax attention scores for a randomly selected head and layer, comparing the exact scores (Exact) with those obtained from simple quantization (Quant, blue). The results show that naive quantization significantly disturbs the distribution shape and introduces excessive outliers in $q K_{\text{deq}}^T/\sqrt{d}$. This justifies the need for an additional correction.

Motivated by these findings, we introduce a post-quantization calibration by re-scaling the pre-softmax attention scores. Specifically, suppose that all elements in $q K_{\text{deq}}^T/\sqrt{d}$ are in the interval $[\gamma, \delta]$. Given $\tau_1,\tau_2>0$, we define a linear transformation $g$ mapping from $[\gamma, \delta]$ to $[\gamma-\tau_1, \delta-\tau_2]$ for $\tau_1,\tau_2>0$, given by:
\begin{align}
g(x) := \frac{\delta-\gamma+\tau_1-\tau_2}{\delta-\gamma}\left( x - \gamma\right) + \gamma - \tau_1.
\end{align}
The attention scores are then adjusted as follows:
\begin{align}
\text{softmax}\left(g \left(\frac{q K_{\text{deq}}^T}{\sqrt{d_k}}\right)\right). \label{eq:attention_scores}
\end{align}
We search for the best calibration parameters $\tau_2, \tau_1$ using a grid search among $\tau_1, \tau_2 \in \{0,1,2,3\}$ and fix it across all prompt inputs, layers, and attention heads. As in \cref{fig-qk-dist}, the calibration (Quant-C, red) effectively mitigates the impact of extreme values, aligning the distribution of the adjusted scores more closely with that of the exact baseline (Exact) compared to the uncalibrated quantization (Quant, blue). Furthermore, as illustrated in \cref{fig-approx-error}, the proposed calibration reduces the mean squared error (MSE) in approximating the attention scores across all layers, outperforming quantization-only approach. Experimental result in \cref{sec-ablation} confirms that this calibration significantly enhances performance on our benchmarks, achieving substantial improvements over the baseline.

\subsection{Implementation Details}






To develop a practical implementation, we consider multi-head attention with $h$ heads. For example, a query can be represented as $q \in \mathbb{R}^{h \times 1 \times d}$. Note that attention computation with quantization requires two key operations: (1) $q K_{\text{dis}}^\top$ and (2) $w  V_{\text{dis}}$ where $K_{\text{dis}}, V_{\text{dis}} \in \{0,1,\dots, 2^b-1\}^{h \times n \times d}$ are discretized KV caches with bitwidth $b$ and $w \in \mathbb{R}^{h \times 1 \times n}$ is the attention scores defined in \cref{eq:attention_scores}.\footnote{The matrix multiplications are performed independently along the first axis, enabling parallel computation.}
Although these multiplications are conceptually straightforward, several obstacles must be addressed to enable an efficient implementation:

\begin{itemize}
\item Lack of hardware intrinsics: While modern GPUs support microscaling formats~\cite{OCP_MicroScaling_2023} to accelerate quantized models down to 4-bit floating points, hardware intrinsics for lower-precision (e.g., 1-bit) encoding remain unavailable.
\item Overhead in matrix multiplication: A naive approach with separate operations fails to reduce host-to-GPU data transfer volumes while introducing additional computational overhead compared to standard matrix multiplication.
\item Irregular matrix shape: Quantization is only partially applied to the visual cache, resulting in computations involving irregular matrix dimensions.
\end{itemize}

To overcome these challenges, we pack the quantized values into the smallest bitwidth (e.g., 8-bit from PyTorch) supported by the GPU. This packing process involves shifting and summing the quantized values to form compact integer representations. More details on packing indices are provided in \cref{sec:packing}.
Additionally, we leverage Triton~\cite{tillet2019triton} to generate optimized kernels that fuse unpacking (decoding) directly into matrix multiplication operations.
For $qK_{\text{dis}}^\top$, the reduction dimension (i.e., $d$) corresponds to compressed input data, which remains compact after unpacking. This compactness enables processing multiple matrices concurrently within each streaming multiprocessor (SM). By doing so, a larger tile of $q$ can be retained in shared memory throughout the computation, reducing costly global memory accesses and improving throughput, as shown in \cref{alg:qk-kernel}. 
For $wV_{\text{dis}}$, the reduction dimension (i.e., $n$) can be large (for long-context input or visual tokens), while the output dimension remains short. This asymmetry allows the kernel to retain a single stripe of $w$ in shared memory throughout the computation, minimizing memory traffic. This constraint favors fine-grained task division across streaming multiprocessors (SMs), where smaller, independent workloads are distributed dynamically to maximize occupancy and hide memory latency. \cref{alg:wv-kernel} demonstrates the workload distribution in the $wV_{\text{dis}}$ kernel.

\begin{algorithm}
\small
\caption{$qK_{\text{dis}}^\top$ Kernel with Fused Unpacking}
\label{alg:qk-kernel}
\SetKwInOut{Input}{Input}
\SetKwInOut{Output}{Output}
\SetKwProg{Proc}{Procedure}{}{}
\Input{$q \in \mathbb{R}^{h \times 1 \times d}$, $K_{\text{dis}} \in \{0,\dots,2^b-1\}^{h \times n \times d_{\text{pack}}}$, \\
block sizes $b_h$, $b_n$
}
\Output{pre-softmax attention scores}
\Proc{\texttt{qk\_kernel}}{
  $\text{pid} \leftarrow \text{program\_id}$ \;
  $\text{start} \leftarrow b_h \times \text{pid}$ \;
  $\text{end} \leftarrow (b_h + 1) \times \text{pid}$ \;
  $\text{q} \leftarrow Q[\text{start}:\text{end}, :, :]$ \tcp*{stays in shared memory}
  \For{$j \leftarrow 0$ \KwTo $n$ \textbf{step} $b_n$}{
    $\text{k} \leftarrow K_{\text{dis}} [\text{start}:\text{end}, j:j+b_n, :]$  \;
    $\text{k}' \leftarrow \texttt{load}(\texttt{unpack}(\text{k})$) \;
    $\text{u} \leftarrow \text{q} \odot \text{k}'$ \tcp*{Element-wise multiplication with boardcast}
    $\text{v} \leftarrow \sum \text{u}$ along axis=1  \;
    $\texttt{store}(\text{v})$
  }
}
\end{algorithm}
\begin{algorithm}
\small
\caption{$wV_{\text{dis}}$ Kernel with Fine-Grained Task Division}
\label{alg:wv-kernel}
\SetKwInOut{Input}{Input}
\SetKwInOut{Output}{Output}
\SetKwProg{Proc}{Procedure}{}{}
\Input{$w \in \mathbb{R}^{h \times 1 \times n}$, $V_{\text{dis}} \in \{0,\dots,2^b-1\}^{h \times n \times d_{\text{pack}}}$,\\
number of all physical SMs $P$, block size $b_h$
}
\Output{attention output with quantized cache}
\Proc{\texttt{wv\_kernel}}{
  $\text{pid} \leftarrow \text{program\_id}$ \;
  $x \leftarrow \frac{d_{\text{pack}}}{b_h}$ \tcp*{Blocks per matrix} 
  $y \leftarrow \left\lceil \frac{h \times x} {P} \right\rceil$ \tcp*{Tasks per SM}
  $s \leftarrow y \times \text{pid}$ \;
  $t \leftarrow y \times (\text{pid} + 1)$ \;
  $i \leftarrow s$ \;
  \While{$i \neq t$}{
    $\text{matrix} \leftarrow \left\lfloor \frac{i}{x} \right\rfloor$ \tcp*{Matrix index}
    $\text{end} \leftarrow \min\big((\text{matrix} + 1) \times b_h,\ t\big)$\;
    $\text{w} \leftarrow w[\text{matrix}, :, :]$ \tcp*{Load $w$ stripe into shared memory}
    \For{$j \leftarrow i$ \KwTo $\text{end}$}{
      $\text{compute output}[\text{matrix}, :, j \times b_h : (j+1) \times b_h]$
    }
    $i \leftarrow \text{end}$ \;
  }
}
\end{algorithm}

\section{Experiment} \label{sec-experiments}

We evaluate the proposed quantization method across multiple tasks where MLLMs demonstrate strong performance. First, we benchmark image captioning task on the COCO Caption dataset~\cite{chen2015microsoft}. Next, we conduct document visual question answering on the DocVQA dataset~\cite{mathew2021docvqa}. We then evaluate video understanding performance on the MMBench-Video~\cite{fang2024mmbenchvideo}. Finally, we report inference speed of our method by measuring throughput. All experiments were performed using NVIDIA H100 GPUs with 80GB VRAM.

\begin{table*}[t]
    \centering
    \scalebox{0.72}{
    \begin{tabular}{lccccccccc}
        \toprule
        \multirow{2}{*}{Methods} & \multirow{2}{*}{Bitwidth $b$} & \multicolumn{8}{c}{Evaluation Metrics ($\uparrow$)} \\
        \cmidrule(lr){3-10}
         & & SPICE & BLEU\_1 & BLEU\_2 & BLEU\_3 & BLEU\_4 & METEOR & ROUGE\_L & CIDEr \\
        \midrule
        \rowcolor[rgb]{ .949,  .949,  .949} \multicolumn{10}{c}{Model: $\mathtt{llava}$-$\mathtt{1.5}$-$\mathtt{7b}$} \\
        Baseline & 16 & 0.235 & 0.731 & 0.563 & 0.413 & 0.295 & 0.292 & 0.558 & 1.105 \\
        VLCache & 8 & 0.233&0.732&0.563&0.413&0.295&0.291&0.556&1.103\\
        Ours & 8 & 0.235&0.731&0.563&0.413&0.295&0.293&0.558&{\bf 1.105} \\
        \midrule
        KIVI & 4 & 0.235&0.730&0.563&0.413&0.296&0.293&0.558& {\bf 1.106} \\
        VLCache & 4 & 0.230 & 0.731& 0.560& 0.410& 0.293 & 0.288 & 0.554 & 1.091 \\
        Ours & 4 & 0.235 & 0.730 & 0.563 & 0.413 & 0.296 & 0.293 & 0.558 & 1.105 \\
        \midrule
        KIVI & 2 & 0.235&0.728&0.560&0.410&0.293&0.292&0.556& {\bf 1.099} \\
        VLCache & 2  & 0.225 & 0.729 & 0.557 & 0.406 & 0.289 & 0.285 & 0.550 & 1.074 \\
        Ours & 2 & 0.235 & 0.729 & 0.561 & 0.412 & 0.295 & 0.292 & 0.557 & {\bf 1.099}\\
        \midrule
        VLCache & 1 & 0.218&0.723&0.551&0.401&0.284&0.281&0.545&1.053 \\
        Ours & 1 & 0.227&0.739&0.571&0.419&0.300&0.287&0.558&{\bf 1.109}\\
        \toprule
        \rowcolor[rgb]{ .949,  .949,  .949} \multicolumn{10}{c}{Model: $\mathtt{llava}$-$\mathtt{1.5}$-$\mathtt{13b}$} \\
        Baseline & 16 & 0.239 & 0.747 & 0.582 & 0.434 & 0.316 & 0.296 & 0.564 & 1.159 \\
        VLCache & 8 & 0.236&0.752&0.585&0.436&0.316&0.294&0.565&{\bf 1.163} \\
        Ours & 8 & 0.239&0.747&0.582&0.434&0.316&0.296&0.565&1.159\\
        \midrule
        KIVI & 4 & 0.240&0.747&0.583&0.435&0.316&0.296&0.565&1.160 \\
        VLCache & 4 & 0.233 & 0.752 & 0.586 & 0.436 & 0.317 & 0.292 & 0.563 & {\bf 1.166} \\
        Ours & 4 & 0.239 & 0.747 & 0.582 & 0.434 & 0.316 & 0.296 & 0.564 & 1.159 \\
        \midrule
        KIVI & 2 & 0.240&0.742&0.578&0.430&0.313&0.296&0.563&1.149 \\
        VLCache & 2 & 0.227&0.753&0.586&0.436&0.316&0.288&0.559&1.150\\
        Ours & 2 & 0.239 & 0.746 & 0.581 & 0.433 & 0.314 & 0.295 & 0.564 & {\bf 1.155}\\
        \midrule
        VLCache & 1 & 0.223&0.751&0.584&0.434&0.314&0.284&0.556&1.137 \\
        Ours & 1 & 0.230 & 0.764 & 0.597 & 0.445 & 0.323 & 0.288 & 0.566 & {\bf 1.168}\\
        \toprule 
        \rowcolor[rgb]{ .949,  .949,  .949} \multicolumn{10}{c}{Model: $\mathtt{internvl}$-$\mathtt{2.5}$-$\mathtt{8b}$} \\
        Baseline & 16 & 0.235 & 0.795 & 0.629 & 0.477 & 0.352 & 0.292 & 0.580 & 1.257\\
        VLCache & 8 & 0.236 & 0.794 & 0.628 & 0.476 & 0.351 & 0.291 & 0.580 & 1.252 \\
        Ours & 8 & 0.236 & 0.795 & 0.630 & 0.476 & 0.351 & 0.292 & 0.580 & {\bf 1.257} \\
        \midrule
        KIVI & 4 & 0.232 & 0.8 & 0.635 & 0.481 & 0.355 & 0.289 & 0.580 & 1.255 \\
        VLCache & 4 & 0.237 & 0.793 & 0.628 & 0.475 & 0.350 & 0.291 & 0.579 & 1.252 \\
        Ours & 4 & 0.236 & 0.795 & 0.630 & 0.477 & 0.352 & 0.292 & 0.580 & {\bf 1.259} \\
        \midrule
        KIVI & 2 & 0.233 & 0.801 & 0.635 & 0.480 & 0.354 & 0.290 & 0.581 & 1.252\\
        VLCache & 2 & 0.235 & 0.793 & 0.628 & 0.474 & 0.349 & 0.290 & 0.577 & 1.250\\
        Ours & 2 & 0.232 & 0.798 & 0.632 & 0.477 & 0.351 & 0.289 & 0.579 & {\bf 1.254}\\
        \midrule
        KIVI & 1 & 0.230 & 0.784 & 0.617 & 0.464 & 0.339 & 0.285 & 0.572 & 1.194 \\
        VLCache & 1 & 0.232 & 0.792 & 0.625 & 0.472 & 0.347 & 0.288 & 0.574 & {\bf 1.236}\\
        Ours & 1 & 0.231 & 0.792 & 0.625 & 0.471 & 0.346 & 0.287 & 0.577 & 1.231 \\
        \toprule
        \rowcolor[rgb]{ .949,  .949,  .949} \multicolumn{10}{c}{Model: $\mathtt{internvl}$-$\mathtt{2.5}$-$\mathtt{26b}$} \\
        Baseline & 16 & 0.244 & 0.813 & 0.653 & 0.499 & 0.374 & 0.300 & 0.594 & 1.321 \\
        VLCache & 8 & 0.242&0.813&0.654&0.501&0.375&0.299&0.593&1.321 \\
        Ours & 8 & 0.244&0.813&0.654&0.499&0.374&0.300&0.593& 1.321 \\
        \midrule
        KIVI & 4 & 0.239 & 0.808 & 0.644 & 0.489 & 0.361 & 0.296 & 0.588 & 1.289\\
        VLCache & 4 & 0.241 & 0.813 & 0.653 & 0.501 & 0.375 & 0.298 & 0.591 & 1.319 \\
        Ours & 4 & 0.243 & 0.813 & 0.654 & 0.500 & 0.374 & 0.300 & 0.594 & {\bf 1.320} \\
        \midrule
        KIVI & 2 & 0.239 & 0.806 & 0.643 & 0.488 & 0.362 & 0.296 & 0.588 & 1.284 \\
        VLCache & 2 & 0.238 & 0.809 & 0.648 & 0.495 & 0.371 & 0.295 & 0.587 & 1.302\\
        Ours & 2 & 0.243 & 0.812 & 0.651 & 0.497 & 0.371 & 0.299 & 0.592 & {\bf 1.313}\\
        \midrule
        KIVI & 1 & 0.237 & 0.794 & 0.631 & 0.479 & 0.355 & 0.292 & 0.579 & 1.261\\
        VLCache & 1 & 0.234 & 0.802 & 0.640 & 0.488 & 0.364 & 0.291 & 0.582 & {\bf 1.282}\\
        Ours & 1 & 0.238 & 0.802 & 0.640 & 0.486 & 0.360 & 0.293 & 0.586 & 1.280\\
        \bottomrule
    \end{tabular}
    }
    \caption{Performance evaluations on COCO Caption~\citep{chen2015microsoft} of various KV cache compression methods for different models. Among all metrics, the CIDEr score is the most conclusive metric for image captioning, aligning closely with human judgment.}
    \label{tab-coco}
\end{table*}


\subsection{Image Captioning} \label{sec-image-captioning}

We test our quantization method on the image captioning task using the COCO Caption dataset~\cite{chen2015microsoft}. We use the following models: 
$\mathtt{llava}$-$\mathtt{1.5}$-$\mathtt{7b}$, $\mathtt{13b}$, 
and $\mathtt{internvl}$-$\mathtt{2.5}$-$\mathtt{8b}$, $\mathtt{26b}$.
An input prompt is constructed with a system prompt and the image tokens andevaluateses the output generation using standard captioning metrics, including BLEU, METEOR, ROUGE-L, SPICE, and CIDEr, to assess both lexical similarity and semantic coherence with ground-truth captions. 
We compare our quantization with other KV cache quantization or compression methods including KIVI~\cite{liukivi} and VLCache~\cite{tu2024vl}. 
For VLCache, we adjust its hyperparameter such as compression ratio so that the total memory budget matches that in others. 
For each model, we evaluate the quality of generations where the bidwidth $b$ of each method is changing from $8$ to $1$.

\cref{tab-coco} summarizes the results. The proposed method (``Ours'') consistently demonstrate competitive or superior performance across different bitwidths (8, 4, 2, and 1 bits) compared to the baseline (16-bit) and other methods such as VLCache and KIVI. In particular, for $\mathtt{llava}$-$\mathtt{1.5}$-$\mathtt{7b}$, our method achieves the highest CIDEr score of 1.105 at 8 bits, matching the baseline, and improved to 1.109 at 1 bit, surpassing VLCache (1.053). Similarly, for $\mathtt{internvl}$-$\mathtt{2.5}$-$\mathtt{26b}$, our method yielded the highest CIDEr score of 1.32 at 4 bits and 1.313 at 2 bits, outperforming both VLCache and KIVI. These results highlight the efficacy of our approach in maintaining or enhancing performance under reduced bitwidths, demonstrating its robustness across diverse model architectures and quantization levels.

\subsection{Document Visual Question Answering}

Next, we evaluate the performance of our method on document visual question answering task using the DocVQA dataset~\cite{mathew2021docvqa} with the $\mathtt{internvl}$-$\mathtt{2.5}$ models. Performance is evaluated using the Average Normalized Levenshtein Similarity (ANLS), where higher values indicate better accuracy. Our proposed method demonstrates robust performance across different bitwidths (4, 2, and 1 bits) compared to the baseline (16-bit) and other competing methods including KIVI and VLCache. Results are reported in \cref{tab-docvqa}. Observe that at 4 bits our method achieves ANLS scores of 0.9138 and 0.9237 for each model respectively, closely matching or slightly exceeding the baseline scores of 0.9135 and 0.9242. At 2 bits, our method maintains competitive performance with ANLS values of 0.8937 and 0.9037, outperforming KIVI (0.8877 and 0.9056) and VLCache (0.8881 and 0.8869). Even at the challenging 1-bit level, our method achieves ANLS scores of 0.8455 and 0.8894, surpassing KIVI (0.8023 and 0.8617) and performing comparably to VLCache (0.8558 and 0.87). These results highlight the efficacy of our approach in preserving accuracy under aggressive quantization, demonstrating its versatility and adaptability across different model sizes on the DocVQA task.

\setlength{\tabcolsep}{10pt}
\begin{table}[th]
    \centering
    \scalebox{0.8}{
    \begin{tabular}{lcccccc}
    \toprule
    Methods & \makebox[1.2cm][c]{Bitwidth $b$} & \multicolumn{4}{c}{ANLS ($\uparrow$)} \\
    \midrule
    \rowcolor[rgb]{ .949,  .949,  .949} & & \multicolumn{2}{c}{$\mathtt{internvl}$-$\mathtt{2.5}$} & \multicolumn{2}{c}
    {$\mathtt{llava}$-$\mathtt{1.5}$} \\
    \rowcolor[rgb]{ .949,  .949,  .949} & & $\mathtt{8b}$ & $\mathtt{26b}$ & $\mathtt{7b}$ & $\mathtt{13b}$ \\
    Baseline & 16 & 0.9135 & 0.9242 & 0.2131 & 0.2368 \\
    KIVI & 8 & 0.9072 & 0.9072 & - & - \\
    VLCache & 8 & 0.9121 & 0.9212 & 0.2055 & 0.2278 \\
    Ours & 8 & {\bf 0.9131} & {\bf 0.9241} & {\bf 0.2131} & {\bf 0.2368} \\
    \midrule
    KIVI & 4 & 0.9074 & 0.9074 & 0.2127 & 0.2367 \\
    VLCache & 4 & 0.9048 & 0.9091 & 0.1955 & 0.2196 \\
    Ours & 4 & {\bf 0.9138} & {\bf 0.9237} & {\bf 0.2133} & {\bf 0.2368} \\
    \midrule
    KIVI & 2 & 0.8877 & 0.9056 & 0.2116 & 0.2379 \\
    VLCache & 2 & 0.8881 & 0.8869 & 0.1865 & 0.2089 \\
    Ours & 2 & {\bf 0.8937} & {\bf 0.9037} & {\bf 0.2133} & {\bf 0.2368} \\
    \midrule
    KIVI & 1 & 0.8023 & 0.8617 & - & - \\
    VLCache & 1 & {\bf 0.8558} & 0.8700 & 0.1783 & 0.1960 \\
    Ours & 1 & 0.8455 & {\bf 0.8894} & {\bf 0.1927} & {\bf 0.2161} \\
    \bottomrule
    \end{tabular}
    }
    \caption{Performance evaluations on DocVQA~\cite{mathew2021docvqa} by Average Normalized Levenshtein Similarity (ANLS) for different methods using LLaVA and InternVL models.}
    \label{tab-docvqa}
\end{table}
\setlength{\tabcolsep}{6pt}

\subsection{Video Understanding}

We benchmark quantization methods on the video understanding task using the MMBench-Video dataset~\cite{fang2024mmbenchvideo}. Table~\ref{tab-mmbench-video} presents the results of different quantization techniques applied to the $\mathtt{internvl}$-$\mathtt{2.5}$ models, evaluating both perception and overall scores. The baseline with 16-bit full-precision achieves the highest scores, serving as an upper bound for comparison. 

Our method consistently outperforms both KIVI and VLCache across all bitwidths. In particular, at 8-bit and 4-bit, it nearly matches the full-precision baseline, demonstrating minimal loss in perception and overall scores. Even at 2-bit, our approach surpasses VLCache and KIVI, preserving better performance. At 1-bit, while performance naturally degrades, our method still outperforms VLCache and KIVI in overall score (0.8894 vs. 0.87 and 0.8617), suggesting improved resilience at extreme quantization.

At 8 bits, our method achieves perception and overall scores of 1.53 and 1.5 for $\mathtt{internvl}$-$\mathtt{2.5}$-$\mathtt{8b}$ model, matching the baseline, and 1.68 for both metrics in 26b model, also aligning with the baseline. At 4 bits, our method outperforms others with scores of 1.54 and 1.51 for 8b model, and 1.69 and 1.68 for 26b model, surpassing VLCache (1.53 and 1.5; 1.68 and 1.67) and KIVI (1.5 and 1.47; 1.66 and 1.65). At 2 bits, our method maintained strong performance, particularly for 26b model, with scores of 1.67 and 1.66, compared to VLCache (1.64 and 1.64) and KIVI (1.64 and 1.63). However, at 1 bit, our method slightly underperforms the VLCache with marginal degradations. 

\setlength{\tabcolsep}{13pt}
\begin{table}[th]
    \centering
    \scalebox{0.8}{
    \begin{tabular}{
        lccccc
        }
    \toprule
    Methods & Bitwidth $b$ & \multicolumn{2}{c}{Perception ($\uparrow$)} & \multicolumn{2}{c}{Overall ($\uparrow$)} \\
    \midrule
    \rowcolor[rgb]{ .949,  .949,  .949} & & \multicolumn{4}{c}{$\mathtt{internvl}$-$\mathtt{2.5}$} \\
    \rowcolor[rgb]{ .949,  .949,  .949} & & $\mathtt{8b}$ & $\mathtt{26b}$ & $\mathtt{8b}$ & $\mathtt{26b}$ \\
    Baseline & 16 & 1.53 & 1.68 & 1.50 & 1.68 \\
    KIVI & 8 & 1.49 & 1.67 & 1.47 & 1.65 \\
    VLCache & 8 & 1.53 & 1.68 & {\bf 1.51} & 1.67 \\
    Ours & 8 & {\bf 1.53} & {\bf 1.68} & 1.50 & {\bf 1.68} \\
    \midrule
    KIVI & 4 & 1.50 & 1.66 & 1.47 & 1.65 \\
    VLCache & 4 & 1.53 & 1.68 & 1.50 & 1.67 \\
    Ours & 4 & {\bf 1.54} & {\bf 1.69} & {\bf 1.51} & {\bf 1.68} \\
    \midrule
    KIVI & 2 & 1.50 & 1.64 & 1.47 & 1.63 \\
    VLCache & 2 & {\bf 1.51} & 1.64 & {\bf 1.49} & 1.64 \\
    Ours & 2 & 1.50 & {\bf 1.67} & 1.47 & {\bf 1.66} \\
    \midrule
    KIVI & 1 & 1.39 & 1.52 & 1.31 & 1.51 \\
    VLCache & 1 & {\bf 1.50} & {\bf 1.65} & {\bf 1.48} & {\bf 1.64} \\
    Ours & 1 & 1.49 & 1.63 & 1.45 & 1.62 \\
    \bottomrule
    \end{tabular}
    }
    \caption{Performance evaluations on MMBench-Video~\cite{fang2024mmbenchvideo} by perception and overall scores for different methods using InternVL models.}
    \label{tab-mmbench-video}
\end{table}
\setlength{\tabcolsep}{6pt}

\subsection{Runtime Analysis} \label{sec-throughput}
\begin{figure}[!htbp]
    \centering
    \hspace{-0.1in}
    \begin{subfigure}{0.24\textwidth}
        \centering
        \includegraphics[width=\textwidth]{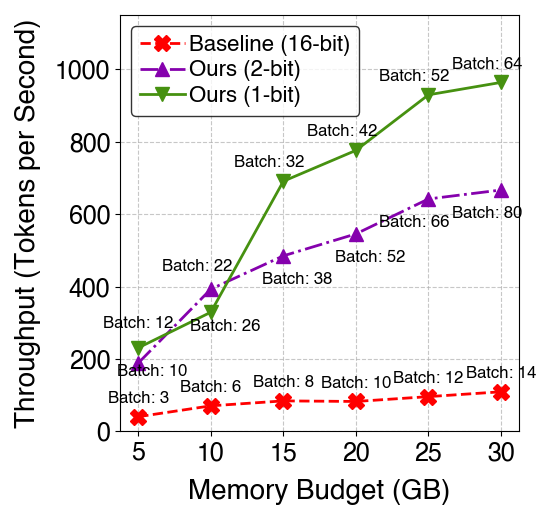}
        \caption{$\mathtt{internvl2\_5}$-$\mathtt{8b}$, $n=3328$}
    \end{subfigure}
    \hspace{-0.1in}
    \begin{subfigure}{0.24\textwidth}
        \centering
        \includegraphics[width=\textwidth]{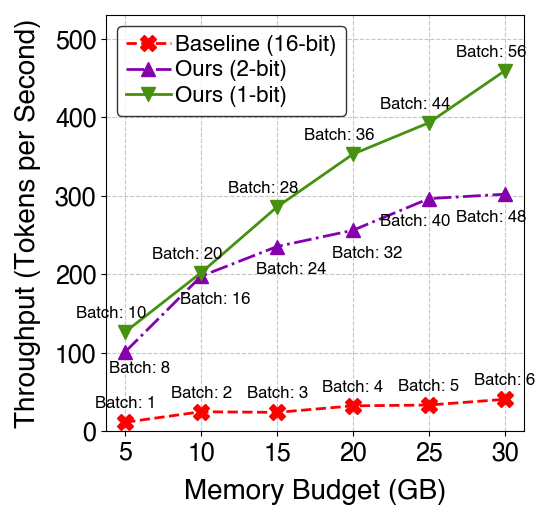}
        \caption{$\mathtt{internvl2\_5}$-$\mathtt{26b}$, $n=3328$}
    \end{subfigure}
    \begin{subfigure}{0.24\textwidth}
        \centering
        \includegraphics[width=\textwidth]{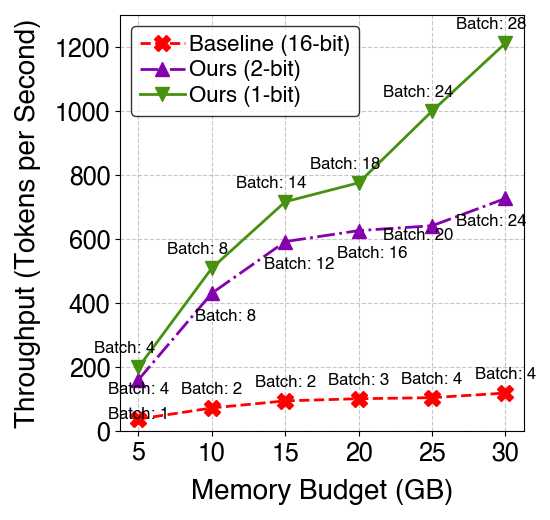}
        \caption{$\mathtt{internvl2\_5}$-$\mathtt{8b}$, $n=8192$}
    \end{subfigure}
    \hspace{-0.1in}
    \begin{subfigure}{0.24\textwidth}
        \centering
        \includegraphics[width=\textwidth]{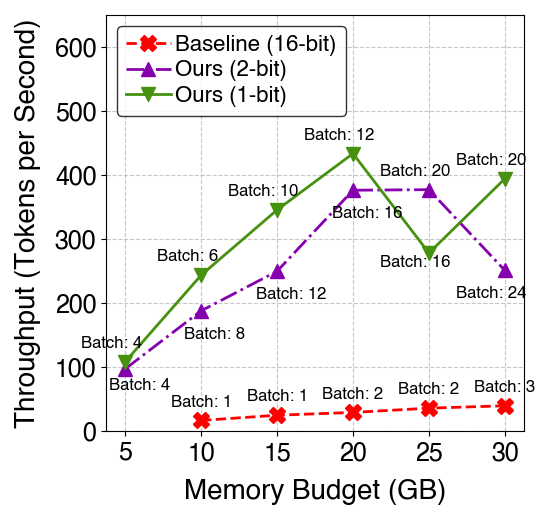}
        \caption{$\mathtt{internvl2\_5}$-$\mathtt{26b}$, $n=8192$}
    \end{subfigure}
    \caption{Throughputs of our 2-bit, 1-bit quantization and the baseline (16-bit) across various memory budgets (5 to 30 GB). We use 2 models: $\mathtt{internvl2\_5}$-$\mathtt{8b}$ and $\mathtt{internvl2\_5}$-$\mathtt{26b}$, and the the visual token lengths are $n=3328$ and $8192$. The annotated texts indicate the maximum batch size accommodated within each memory budget. 
    }
    \label{fig-throughput}
\end{figure}

To show the impact of our quantization on decoding efficiency, we evaluated the throughput (i.e., the number of generated tokens per second) of the proposed 1-bit quantization method against a 16-bit baseline using the  $\mathtt{internvl}$-$\mathtt{2.5}$ models. We consider two scenarios where the lengths of the visual tokens are $n=3328$ and $8192$. We vary the maximum GPU memory from 5 GB to 30 GB and for each memory constraint we find the maximum number of batch size to fit in and measure throughput of the decoding stage. \cref{fig-throughput} demonstrates that our 1-bit quantization method consistently outperforms the baseline in all memory budgets. For example, when $n=3329$ for 8B parameter model, we achieve 126.582 tokens/s at 5 GB (versus 11.628 tokens/s for the baseline) and it scales to 459.016 tokens/s at 30 GB (versus 40.816 tokens/s for the baseline). This represents a throughput boost of approximately $9.88\times$ to $11.24\times$ over the baseline, showing the efficacy of our approach in enhancing decoding performance under constrained memory conditions. We have attached detailed throughput data as an appendix to the paper.


\subsection{Ablation Study} \label{sec-ablation} 

We conduct two ablation studies to validate our key contributions: the calibration technique and channel-wise quantization applied to the value cache. We replicate the image captioning task outlined in \cref{sec-image-captioning} using $\mathtt{internvl}$-$\mathtt{2.5}$-$\mathtt{8b}$ model.

\paragraph{Calibration of Post-quantization.} 
In order to investigate the impact of calibration on pre-softmax attention scores discussed in \cref{sec-calibration}, we fix all hyperparameters of quantization with $b=1$
and compare evaluation metrics of quantizations with and without calibration. As presented \cref{tab-ablation}, calibration (Quant-C) substantially outperforms the uncalibration (Quant) for all evaluation metrics. This justifies the crucial role of calibration to achieve promising performances.

\paragraph{Channel-wise Quantization on Value Cache.}
We additionally compare different quantization approaches for the value cache. In particular, we compare channel-wise to the non-channel-wise, which finds the global minimum and maximum values. In \cref{tab-ablation}, we observe that non-channel-wise quantization performs worse than the channel-wise one. This supports our approach for the value cache. 



\begin{table}[th]
    \centering
    \scalebox{0.85}{
    \begin{tabular}{lccccccccc}
        \toprule
        Metrics ($\uparrow$) & \makecell[c]{Channel-wise \\+ Calibration} & \makecell[c]{Without \\ Calibration} & \makecell[c]{Without \\ Channel-wise} \\
        \midrule
        SPICE & 0.231 &0.230 & 0.235  \\
        BLEU\_1 & 0.792 & 0.784 & 0.792 \\
        BLEU\_2 &  0.625 & 0.617 & 0.609 \\
        BLEU\_3 & 0.471 & 0.464 &  0.457\\
        BLEU\_4 & 0.346 & 0.339 &  0.334 \\
        METEOR & 0.287 & 0.285 &  0.288 \\
        ROUGE\_L & 0.577 & 0.572 &  0.571\\
        CIDEr & {\bf 1.231} &1.194  & 1.200\\
        \bottomrule
    \end{tabular}
    }
    \caption{Comparison of calibration on pre-softmax attentions and channel-wise quantization on the value cache on the COCO Caption dataset.}
    \label{tab-ablation}
\end{table}




\section{Conclusion}
In this paper, we explore the compression of visual caches in multimodal large language models (MLLMs). Unlike prior work that focuses on token dropping, we investigate quantization techniques specifically designed for visual tokens, enabling lower-bit representations. A na\"ive quantization to extreme bit levels often induces distribution shifts, leading to degraded performance. To address this, we propose a novel calibration strategy for pre-softmax attention scores, mitigating quantization-induced distortions. Additionally, we introduce a post-scaling technique for efficient channel-wise cache quantization. 
Experiments on the InternVL model family for COCO Caption, MMBench-Video, and DocVQA benchmarks demonstrate the effectiveness of our approach.

{
    \small
    \bibliographystyle{ieeenat_fullname}
    \bibliography{main}

\begin{thebibliography}{36}
\providecommand{\natexlab}[1]{#1}
\providecommand{\url}[1]{\texttt{#1}}
\expandafter\ifx\csname urlstyle\endcsname\relax
  \providecommand{\doi}[1]{doi: #1}\else
  \providecommand{\doi}{doi: \begingroup \urlstyle{rm}\Url}\fi

\bibitem[Abdin et~al.(2024)Abdin, Aneja, Awadalla, Awadallah, Awan, Bach, Bahree, Bakhtiari, Bao, Behl, et~al.]{abdin2024phi}
Marah Abdin, Jyoti Aneja, Hany Awadalla, Ahmed Awadallah, Ammar~Ahmad Awan, Nguyen Bach, Amit Bahree, Arash Bakhtiari, Jianmin Bao, Harkirat Behl, et~al.
\newblock Phi-3 technical report: A highly capable language model locally on your phone.
\newblock \emph{arXiv preprint arXiv:2404.14219}, 2024.

\bibitem[Bahdanau et~al.(2014)Bahdanau, Cho, and Bengio]{bahdanau2014neural}
Dzmitry Bahdanau, Kyunghyun Cho, and Yoshua Bengio.
\newblock Neural machine translation by jointly learning to align and translate.
\newblock \emph{arXiv preprint arXiv:1409.0473}, 2014.

\bibitem[Bai et~al.(2023)Bai, Bai, Yang, Wang, Tan, Wang, Lin, Zhou, and Zhou]{bai2023qwen}
Jinze Bai, Shuai Bai, Shusheng Yang, Shijie Wang, Sinan Tan, Peng Wang, Junyang Lin, Chang Zhou, and Jingren Zhou.
\newblock Qwen-vl: A frontier large vision-language model with versatile abilities.
\newblock \emph{arXiv preprint arXiv:2308.12966}, 2023.

\bibitem[Cai et~al.(2024)Cai, Zhang, Gao, Liu, Liu, Lu, Xiong, Dong, Chang, Hu, et~al.]{cai2024pyramidkv}
Zefan Cai, Yichi Zhang, Bofei Gao, Yuliang Liu, Tianyu Liu, Keming Lu, Wayne Xiong, Yue Dong, Baobao Chang, Junjie Hu, et~al.
\newblock Pyramidkv: Dynamic kv cache compression based on pyramidal information funneling.
\newblock \emph{arXiv preprint arXiv:2406.02069}, 2024.

\bibitem[Chen et~al.(2024{\natexlab{a}})Chen, Chen, Zhang, Liu, Wang, Zhou, Zhang, Wan, Zhou, and Sun]{chen2024mllm}
Dongping Chen, Ruoxi Chen, Shilin Zhang, Yinuo Liu, Yaochen Wang, Huichi Zhou, Qihui Zhang, Yao Wan, Pan Zhou, and Lichao Sun.
\newblock Mllm-as-a-judge: Assessing multimodal llm-as-a-judge with vision-language benchmark.
\newblock \emph{arXiv preprint arXiv:2402.04788}, 2024{\natexlab{a}}.

\bibitem[Chen et~al.(2024{\natexlab{b}})Chen, Zhao, Liu, Bai, Lin, Zhou, and Chang]{chen2024fastv}
Liang Chen, Haozhe Zhao, Tianyu Liu, Shuai Bai, Junyang Lin, Chang Zhou, and Baobao Chang.
\newblock An image is worth 1/2 tokens after layer 2: Plug-and-play inference acceleration for large vision-language models.
\newblock In \emph{European Conference on Computer Vision}, pages 19--35. Springer, 2024{\natexlab{b}}.

\bibitem[Chen et~al.(2015)Chen, Fang, Lin, Vedantam, Gupta, Doll{\'a}r, and Zitnick]{chen2015microsoft}
Xinlei Chen, Hao Fang, Tsung-Yi Lin, Ramakrishna Vedantam, Saurabh Gupta, Piotr Doll{\'a}r, and C~Lawrence Zitnick.
\newblock Microsoft coco captions: Data collection and evaluation server.
\newblock \emph{arXiv preprint arXiv:1504.00325}, 2015.

\bibitem[Chu et~al.(2023)Chu, Qiao, Lin, Xu, Yang, Hu, Wei, Zhang, Zhang, Wei, et~al.]{chu2023mobilevlm}
Xiangxiang Chu, Limeng Qiao, Xinyang Lin, Shuang Xu, Yang Yang, Yiming Hu, Fei Wei, Xinyu Zhang, Bo Zhang, Xiaolin Wei, et~al.
\newblock Mobilevlm: A fast, strong and open vision language assistant for mobile devices.
\newblock \emph{arXiv preprint arXiv:2312.16886}, 2023.

\bibitem[Chu et~al.(2024)Chu, Qiao, Zhang, Xu, Wei, Yang, Sun, Hu, Lin, Zhang, et~al.]{chu2024mobilevlm}
Xiangxiang Chu, Limeng Qiao, Xinyu Zhang, Shuang Xu, Fei Wei, Yang Yang, Xiaofei Sun, Yiming Hu, Xinyang Lin, Bo Zhang, et~al.
\newblock Mobilevlm v2: Faster and stronger baseline for vision language model.
\newblock \emph{arXiv preprint arXiv:2402.03766}, 2024.

\bibitem[Dai et~al.(2024)Dai, Deng, Zhao, Xu, Gao, Chen, Li, Zeng, Yu, Wu, et~al.]{dai2024deepseekmoe}
Damai Dai, Chengqi Deng, Chenggang Zhao, RX Xu, Huazuo Gao, Deli Chen, Jiashi Li, Wangding Zeng, Xingkai Yu, Y Wu, et~al.
\newblock Deepseekmoe: Towards ultimate expert specialization in mixture-of-experts language models.
\newblock \emph{arXiv preprint arXiv:2401.06066}, 2024.

\bibitem[Fang et~al.(2024)Fang, Mao, Duan, Zhao, Li, Lin, and Chen]{fang2024mmbenchvideo}
Xinyu Fang, Kangrui Mao, Haodong Duan, Xiangyu Zhao, Yining Li, Dahua Lin, and Kai Chen.
\newblock Mmbench-video: A long-form multi-shot benchmark for holistic video understanding.
\newblock \emph{arXiv preprint arXiv:2406.14515}, 2024.

\bibitem[Han et~al.(2025{\natexlab{a}})Han, Kacham, Karbasi, Mirrokni, and Zandieh]{han2025polarquant}
Insu Han, Praneeth Kacham, Amin Karbasi, Vahab Mirrokni, and Amir Zandieh.
\newblock Polarquant: Quantizing kv caches with polar transformation.
\newblock \emph{arXiv preprint arXiv:2502.02617}, 2025{\natexlab{a}}.

\bibitem[Han et~al.(2025{\natexlab{b}})Han, Kapralov, Kochetkova, Sheth, and Zandieh]{han2025balancekv}
Insu Han, Michael Kapralov, Ekaterina Kochetkova, Kshiteej Sheth, and Amir Zandieh.
\newblock Balancekv: Kv cache compression through discrepancy theory.
\newblock \emph{arXiv preprint arXiv:2502.07861}, 2025{\natexlab{b}}.

\bibitem[Hooper et~al.(2024)Hooper, Kim, Mohammadzadeh, Mahoney, Shao, Keutzer, and Gholami]{hooper2024kvquant}
Coleman Hooper, Sehoon Kim, Hiva Mohammadzadeh, Michael~W Mahoney, Sophia Shao, Kurt Keutzer, and Amir Gholami.
\newblock Kvquant: Towards 10 million context length llm inference with kv cache quantization.
\newblock \emph{Advances in Neural Information Processing Systems}, 37:\penalty0 1270--1303, 2024.

\bibitem[Jiang et~al.(2024)Jiang, Sablayrolles, Roux, Mensch, Savary, Bamford, Chaplot, Casas, Hanna, Bressand, et~al.]{jiang2024mixtral}
Albert~Q Jiang, Alexandre Sablayrolles, Antoine Roux, Arthur Mensch, Blanche Savary, Chris Bamford, Devendra~Singh Chaplot, Diego de~las Casas, Emma~Bou Hanna, Florian Bressand, et~al.
\newblock Mixtral of experts.
\newblock \emph{arXiv preprint arXiv:2401.04088}, 2024.

\bibitem[Jin et~al.(2024)Jin, Han, Yang, Jiang, Liu, Chang, Chen, and Hu]{jin2024llm}
Hongye Jin, Xiaotian Han, Jingfeng Yang, Zhimeng Jiang, Zirui Liu, Chia-Yuan Chang, Huiyuan Chen, and Xia Hu.
\newblock Llm maybe longlm: Self-extend llm context window without tuning.
\newblock \emph{arXiv preprint arXiv:2401.01325}, 2024.

\bibitem[Kwon et~al.(2023)Kwon, Li, Zhuang, Sheng, Zheng, Yu, Gonzalez, Zhang, and Stoica]{kwon2023efficient}
Woosuk Kwon, Zhuohan Li, Siyuan Zhuang, Ying Sheng, Lianmin Zheng, Cody~Hao Yu, Joseph Gonzalez, Hao Zhang, and Ion Stoica.
\newblock Efficient memory management for large language model serving with pagedattention.
\newblock In \emph{Proceedings of the 29th Symposium on Operating Systems Principles}, pages 611--626, 2023.

\bibitem[Li et~al.(2024)Li, Zhang, Wang, Zhong, Chen, Chu, Liu, and Jia]{li2024mini}
Yanwei Li, Yuechen Zhang, Chengyao Wang, Zhisheng Zhong, Yixin Chen, Ruihang Chu, Shaoteng Liu, and Jiaya Jia.
\newblock Mini-gemini: Mining the potential of multi-modality vision language models.
\newblock \emph{arXiv preprint arXiv:2403.18814}, 2024.

\bibitem[Lin et~al.(2024)Lin, Tang, Ye, Cui, Zhu, Jin, Huang, Zhang, Pang, Ning, et~al.]{lin2024moe}
Bin Lin, Zhenyu Tang, Yang Ye, Jiaxi Cui, Bin Zhu, Peng Jin, Jinfa Huang, Junwu Zhang, Yatian Pang, Munan Ning, et~al.
\newblock Moe-llava: Mixture of experts for large vision-language models.
\newblock \emph{arXiv preprint arXiv:2401.15947}, 2024.

\bibitem[Liu et~al.(2024)Liu, Yuan, Jin, Zhong, Xu, Braverman, Chen, and Hu]{liukivi}
Zirui Liu, Jiayi Yuan, Hongye Jin, Shaochen Zhong, Zhaozhuo Xu, Vladimir Braverman, Beidi Chen, and Xia Hu.
\newblock Kivi: A tuning-free asymmetric 2bit quantization for kv cache.
\newblock In \emph{International Conference on Machine Learning}, 2024.

\bibitem[Mathew et~al.(2021)Mathew, Karatzas, and Jawahar]{mathew2021docvqa}
Minesh Mathew, Dimosthenis Karatzas, and CV Jawahar.
\newblock Docvqa: A dataset for vqa on document images.
\newblock In \emph{Proceedings of the IEEE/CVF winter conference on applications of computer vision}, pages 2200--2209, 2021.

\bibitem[Rouhani et~al.(2023)Rouhani, Garegrat, Savell, More, Han, Zhao, Hall, Klar, Chung, Yu, Schulte, Wittig, Bratt, Stephens, Milanovic, Brothers, Dubey, Cornea, Heinecke, Rodriguez, Langhammer, Deng, Naumov, Micikevicius, Siu, and Verrilli]{OCP_MicroScaling_2023}
Bita~Darvish Rouhani, Nitin Garegrat, Tom Savell, Ankit More, Kyung-Nam Han, Ritchie Zhao, Mathew Hall, Jasmine Klar, Eric Chung, Yuan Yu, Michael Schulte, Ralph Wittig, Ian Bratt, Nigel Stephens, Jelena Milanovic, John Brothers, Pradeep Dubey, Marius Cornea, Alexander Heinecke, Andres Rodriguez, Martin Langhammer, Summer Deng, Maxim Naumov, Paulius Micikevicius, Michael Siu, and Colin Verrilli.
\newblock {OCP} micro scaling formats mx v1.0 specification.
\newblock Technical report, Open Compute Project, 2023.
\newblock Version 1.0.

\bibitem[Shang et~al.(2024)Shang, Cai, Xu, Lee, and Yan]{shang2024llava-prumerge}
Yuzhang Shang, Mu Cai, Bingxin Xu, Yong~Jae Lee, and Yan Yan.
\newblock Llava-prumerge: Adaptive token reduction for efficient large multimodal models.
\newblock \emph{arXiv preprint arXiv:2403.15388}, 2024.

\bibitem[Tang et~al.(2023)Tang, Bi, Xu, Song, Liang, Wang, Zhang, An, Lin, Zhu, et~al.]{tang2023video}
Yunlong Tang, Jing Bi, Siting Xu, Luchuan Song, Susan Liang, Teng Wang, Daoan Zhang, Jie An, Jingyang Lin, Rongyi Zhu, et~al.
\newblock Video understanding with large language models: A survey.
\newblock \emph{arXiv preprint arXiv:2312.17432}, 2023.

\bibitem[Tillet et~al.(2019)Tillet, Kung, and Cox]{tillet2019triton}
Philippe Tillet, H.~T. Kung, and David Cox.
\newblock Triton: an intermediate language and compiler for tiled neural network computations.
\newblock In \emph{Proceedings of the 3rd ACM SIGPLAN International Workshop on Machine Learning and Programming Languages}, page 10–19, New York, NY, USA, 2019. Association for Computing Machinery.

\bibitem[Tu et~al.(2024)Tu, Vashchilenko, Lu, and Xu]{tu2024vl}
Dezhan Tu, Danylo Vashchilenko, Yuzhe Lu, and Panpan Xu.
\newblock Vl-cache: Sparsity and modality-aware kv cache compression for vision-language model inference acceleration.
\newblock \emph{arXiv preprint arXiv:2410.23317}, 2024.

\bibitem[Vaswani(2017)]{vaswani2017attention}
A Vaswani.
\newblock Attention is all you need.
\newblock \emph{Advances in Neural Information Processing Systems}, 2017.

\bibitem[Wan et~al.(2024)Wan, Wu, Liu, Huang, Zhu, Jin, Wang, and Yuan]{wan2024look}
Zhongwei Wan, Ziang Wu, Che Liu, Jinfa Huang, Zhihong Zhu, Peng Jin, Longyue Wang, and Li Yuan.
\newblock Look-m: Look-once optimization in kv cache for efficient multimodal long-context inference.
\newblock \emph{arXiv preprint arXiv:2406.18139}, 2024.

\bibitem[Wu et~al.(2020)Wu, Judd, Zhang, Isaev, and Micikevicius]{wu2020integer}
Hao Wu, Patrick Judd, Xiaojie Zhang, Mikhail Isaev, and Paulius Micikevicius.
\newblock Integer quantization for deep learning inference: Principles and empirical evaluation.
\newblock \emph{arXiv preprint arXiv:2004.09602}, 2020.

\bibitem[Xing et~al.(2024)Xing, Huang, Dong, Lu, Zhang, Zang, Cao, He, Wang, Wu, et~al.]{xing2024pyramiddrop}
Long Xing, Qidong Huang, Xiaoyi Dong, Jiajie Lu, Pan Zhang, Yuhang Zang, Yuhang Cao, Conghui He, Jiaqi Wang, Feng Wu, et~al.
\newblock Pyramiddrop: Accelerating your large vision-language models via pyramid visual redundancy reduction.
\newblock \emph{arXiv preprint arXiv:2410.17247}, 2024.

\bibitem[Yao et~al.(2024)Yao, Yu, Zhang, Wang, Cui, Zhu, Cai, Li, Zhao, He, et~al.]{yao2024minicpm}
Yuan Yao, Tianyu Yu, Ao Zhang, Chongyi Wang, Junbo Cui, Hongji Zhu, Tianchi Cai, Haoyu Li, Weilin Zhao, Zhihui He, et~al.
\newblock Minicpm-v: A gpt-4v level mllm on your phone.
\newblock \emph{arXiv preprint arXiv:2408.01800}, 2024.

\bibitem[Zandieh et~al.(2024{\natexlab{a}})Zandieh, Daliri, and Han]{zandieh2024qjl}
Amir Zandieh, Majid Daliri, and Insu Han.
\newblock Qjl: 1-bit quantized jl transform for kv cache quantization with zero overhead.
\newblock \emph{arXiv preprint arXiv:2406.03482}, 2024{\natexlab{a}}.

\bibitem[Zandieh et~al.(2024{\natexlab{b}})Zandieh, Han, Mirrokni, and Karbasi]{zandieh2024subgen}
Amir Zandieh, Insu Han, Vahab Mirrokni, and Amin Karbasi.
\newblock Subgen: Token generation in sublinear time and memory.
\newblock \emph{arXiv preprint arXiv:2402.06082}, 2024{\natexlab{b}}.

\bibitem[Zhang et~al.(2023)Zhang, Sheng, Zhou, Chen, Zheng, Cai, Song, Tian, R{\'e}, Barrett, et~al.]{zhang2023h2o}
Zhenyu Zhang, Ying Sheng, Tianyi Zhou, Tianlong Chen, Lianmin Zheng, Ruisi Cai, Zhao Song, Yuandong Tian, Christopher R{\'e}, Clark Barrett, et~al.
\newblock H2o: Heavy-hitter oracle for efficient generative inference of large language models.
\newblock \emph{Advances in Neural Information Processing Systems}, 36:\penalty0 34661--34710, 2023.

\bibitem[Zhang et~al.(2024{\natexlab{a}})Zhang, Liu, Cheng, Xu, and Gao]{zhang2024diversifying}
Zeliang Zhang, Xiaodong Liu, Hao Cheng, Chenliang Xu, and Jianfeng Gao.
\newblock Diversifying the expert knowledge for task-agnostic pruning in sparse mixture-of-experts.
\newblock \emph{arXiv preprint arXiv:2407.09590}, 2024{\natexlab{a}}.

\bibitem[Zhang et~al.(2024{\natexlab{b}})Zhang, Pham, Zhao, Wan, Li, Zhou, Miranda, Kale, and Xu]{zhang2024treat}
Zeliang Zhang, Phu Pham, Wentian Zhao, Kun Wan, Yu-Jhe Li, Jianing Zhou, Daniel Miranda, Ajinkya Kale, and Chenliang Xu.
\newblock Treat visual tokens as text? but your mllm only needs fewer efforts to see.
\newblock \emph{arXiv preprint arXiv:2410.06169}, 2024{\natexlab{b}}.

\end{thebibliography}
}

\clearpage
\appendix
\renewcommand{\appendixpagename}{Appendix}
\onecolumn

\appendixpage

\section{Throughput Analysis}

We analyzed the throughputs on InternVL models with different token lengths on H100. The results are shown in the following tables. \cref{tbl:vl8-3328} and \cref{tbl:vl26-3328} covers the results with sequence length $3328$ and parameters $8B$ and $26B$ respectively. \cref{tbl:vl8-8192} and \cref{tbl:vl26-8192} covers the results with larger sequence length ($8192$) and the same parameter configurations.

\begin{table*}[htbp]
    \centering
    \resizebox{\linewidth}{!}{
    \begin{tabular}{llcccccc}
        \toprule
        \textbf{Model} & \textbf{Memory (GB)} & \textbf{Max BS} & \textbf{Prefill Speed} & \textbf{Decode Speed (s/token)} & \textbf{Throughput Decode} & \textbf{Throughput Overall (500 tokens)} \\
        \midrule
        \texttt{internvl-8b-16bit} & 5  & 3  & 0.117 & 0.074 & 40.54  & 0.0808 \\
                                   & 10 & 6  & 0.176 & 0.085 & 70.59  & 0.1406 \\
                                   & 15 & 8  & 0.212 & 0.095 & 84.21  & 0.1677 \\
                                   & 20 & 10 & 0.253 & 0.121 & 82.64  & 0.1646 \\
                                   & 25 & 12 & 0.293 & 0.125 & 96.00  & 0.1911 \\
                                   & 30 & 14 & 0.365 & 0.128 & 109.38 & 0.2175 \\
        \midrule
        \texttt{internvl-8b-ours-1bit} & 5  & 12  & 0.362 & 0.052 & 230.77 & 0.4552 \\
                                       & 10 & 26  & 0.734 & 0.079 & 329.11 & 0.6462 \\
                                       & 15 & 38  & 1.046 & 0.055 & 690.91 & 1.3312 \\
                                       & 20 & 52  & 1.510 & 0.067 & 776.12 & 1.4853 \\
                                       & 25 & 66  & 1.868 & 0.071 & 929.58 & 1.7662 \\
                                       & 30 & 80  & 2.225 & 0.083 & 963.86 & 1.8296 \\
        \midrule
        \texttt{internvl-8b-ours-2bit} & 5  & 10  & 0.294 & 0.053 & 188.68 & 0.3732 \\
                                       & 10 & 22  & 0.623 & 0.056 & 392.86 & 0.7686 \\
                                       & 15 & 32  & 0.957 & 0.066 & 484.85 & 0.9424 \\
                                       & 20 & 42  & 1.156 & 0.077 & 545.45 & 1.0591 \\
                                       & 25 & 52  & 1.515 & 0.081 & 641.98 & 1.2377 \\
                                       & 30 & 64  & 1.813 & 0.096 & 666.67 & 1.2848 \\
        \bottomrule
    \end{tabular}
    }
    \caption{Throughput of our method on InternVL-8B model with varying memory budgets on H100 with sequence length 3328}
    \label{tbl:vl8-3328}
\end{table*}

\begin{table*}[tbh]
    \centering
     \resizebox{\linewidth}{!}{
    \begin{tabular}{llcccccc}
        \toprule
        \textbf{Model} & \textbf{Memory (GB)} & \textbf{Max BS} & \textbf{Prefill Speed} & \textbf{Decode Speed (s/token)} & \textbf{Throughput Decode} & \textbf{Throughput Overall (500 tokens)} \\
        \midrule
        \texttt{internvl-26b-16bit} & 5  & 1  & 0.067 & 0.086 & 11.63  & 0.0232 \\
                                    & 10 & 2  & 0.116 & 0.080 & 25.00  & 0.0499 \\
                                    & 15 & 3  & 0.167 & 0.124 & 24.19  & 0.0483 \\
                                    & 20 & 4  & 0.210 & 0.123 & 32.52  & 0.0648 \\
                                    & 25 & 5  & 0.275 & 0.149 & 33.56  & 0.0669 \\
                                    & 30 & 6  & 0.319 & 0.147 & 40.82  & 0.0813 \\
        \midrule
        \texttt{internvl-26b-ours-1bit} & 5  & 10  & 0.703 & 0.079 & 126.58 & 0.2487 \\
                                        & 10 & 20  & 1.181 & 0.099 & 202.02 & 0.3946 \\
                                        & 15 & 28  & 1.772 & 0.098 & 285.71 & 0.5515 \\
                                        & 20 & 36  & 2.246 & 0.102 & 352.94 & 0.6761 \\
                                        & 25 & 44  & 2.951 & 0.112 & 392.86 & 0.7464 \\
                                        & 30 & 56  & 3.800 & 0.122 & 459.02 & 0.8642 \\
        \midrule
        \texttt{internvl-26b-ours-2bit} & 5  & 8  & 0.591 & 0.079 & 101.27 & 0.1995 \\
                                        & 10 & 16  & 1.216 & 0.081 & 197.53 & 0.3835 \\
                                        & 15 & 24  & 1.553 & 0.102 & 235.29 & 0.4567 \\
                                        & 20 & 32  & 2.199 & 0.125 & 256.00 & 0.4946 \\
                                        & 25 & 40  & 2.619 & 0.135 & 296.30 & 0.5705 \\
                                        & 30 & 48  & 3.051 & 0.159 & 301.89 & 0.5815 \\
        \bottomrule
    \end{tabular}
    }
    \caption{Throughput of our method on InternVL-26B model with varying memory budgets on H100 with sequence length 3328}
    \label{tbl:vl26-3328}
\end{table*}

\begin{table*}[htbp]
    \centering
    \resizebox{\linewidth}{!}{
    \begin{tabular}{llcccccc}
        \toprule
        \textbf{Model} & \textbf{Memory (GB)} & \textbf{Max BS} & \textbf{Prefill Speed} & \textbf{Decode Speed (s/token)} & \textbf{Throughput Decode} & \textbf{Throughput Overall (500 tokens)} \\
        \midrule
        \texttt{internvl-8b-16bit} & 5  & 1  & 0.115 & 0.078 & 38.46  & 0.0767 \\
                                   & 10 & 2  & 0.177 & 0.082 & 73.17  & 0.1457 \\
                                   & 15 & 2  & 0.176 & 0.084 & 95.24  & 0.1897 \\
                                   & 20 & 3  & 0.236 & 0.098 & 102.04 & 0.2031 \\
                                   & 25 & 4  & 0.294 & 0.114 & 105.26 & 0.2094 \\
                                   & 30 & 4  & 0.294 & 0.117 & 119.66 & 0.2381 \\
        \midrule
        \texttt{internvl-8b-ours-1bit} & 5  & 4  & 0.299 & 0.060 & 200.00 & 0.3961 \\
                                       & 10 & 8  & 0.581 & 0.051 & 509.80 & 0.9969 \\
                                       & 15 & 14 & 0.941 & 0.053 & 716.98 & 1.3848 \\
                                       & 20 & 18 & 1.210 & 0.067 & 776.12 & 1.4981 \\
                                       & 25 & 24 & 1.590 & 0.066 & 1000.00 & 1.9081 \\
                                       & 30 & 28 & 3.110 & 0.066 & 1212.12 & 2.2155 \\
        \midrule
        \texttt{internvl-8b-ours-2bit} & 5  & 4  & 0.300 & 0.062 & 161.29 & 0.3195 \\
                                       & 10 & 8  & 0.584 & 0.051 & 431.37 & 0.8434 \\
                                       & 15 & 12 & 0.813 & 0.054 & 592.59 & 1.1505 \\
                                       & 20 & 16 & 1.074 & 0.067 & 626.87 & 1.2148 \\
                                       & 25 & 20 & 1.358 & 0.081 & 641.98 & 1.2423 \\
                                       & 30 & 24 & 1.598 & 0.088 & 727.27 & 1.4036 \\
        \bottomrule
    \end{tabular}
    }
    \caption{Throughput of our method on InternVL-8B model with varying memory budgets on H100 with sequence length 8192}
    \label{tbl:vl8-8192}
\end{table*}

\begin{table*}[th]
    \centering
    \resizebox{\linewidth}{!}{
    \begin{tabular}{llcccccc}
        \toprule
        \textbf{Model} & \textbf{Memory (GB)} & \textbf{Max BS} & \textbf{Prefill Speed} & \textbf{Decode Speed (s/token)} & \textbf{Throughput Decode} & \textbf{Throughput Overall (500 tokens)} \\
        \midrule
        \texttt{internvl-26b-16bit} & 5  & <1 & --   & --   & --     & --      \\
                                    & 10 & 1  & 0.136 & 0.117 & 17.09  & 0.0341  \\
                                    & 15 & 1  & 0.136 & 0.119 & 25.21  & 0.0503  \\
                                    & 20 & 2  & 0.263 & 0.135 & 29.63  & 0.0590  \\
                                    & 25 & 2  & 0.263 & 0.138 & 36.23  & 0.0722  \\
                                    & 30 & 3  & 0.381 & 0.150 & 40.00  & 0.0796  \\
        \midrule
        \texttt{internvl-26b-ours-1bit} & 5  & 4  & 0.691  & 0.092  & 108.70 & 0.2142 \\
                                        & 10 & 6  & 1.069  & 0.082  & 243.90 & 0.4754 \\
                                        & 15 & 10 & 1.560  & 0.081  & 345.68 & 0.6657 \\
                                        & 20 & 12 & 1.848  & 0.083  & 433.73 & 0.8305 \\
                                        & 25 & 16 & 2.540  & 0.158  & 278.48 & 0.5396 \\
                                        & 30 & 20 & 3.103  & 0.142  & 394.37 & 0.7557 \\
        \midrule
        \texttt{internvl-26b-ours-2bit} & 5  & 2  & 0.358  & 0.082  & 97.56  & 0.1934 \\
                                        & 10 & 4  & 0.677  & 0.085  & 188.24 & 0.3706 \\
                                        & 15 & 6  & 0.943  & 0.096  & 250.00 & 0.4904 \\
                                        & 20 & 10 & 1.574  & 0.085  & 376.47 & 0.7261 \\
                                        & 25 & 12 & 1.857  & 0.106  & 377.36 & 0.7292 \\
                                        & 30 & 16 & 2.550  & 0.191  & 251.31 & 0.4895 \\
        \bottomrule
    \end{tabular}
    }
    \caption{Throughput of our method on InternVL-26B model with varying memory budgets on H100 with sequence length 8192}
    \label{tbl:vl26-8192}
\end{table*}

\subsection{Details for Low-bit Quantization with Packing} \label{sec:packing}

Suppose we aim to quantize data into $N$-bit values and pack them into an $M$-bit integer, where $M$ must be divisible by $N$. Along the quantization dimension (which is contiguous in memory), we partition the data into groups of size $\frac{M}{N}$. For each $i$-th element (starting from $0$) within a group $v$, we left-shift the corresponding quantized value $v[i]$ by $M-N(1+i)$ bits. The packed integer is obtained by summing these shifted values. Since the shifted bits occupy non-overlapping positions, this summation is equivalent to a bitwise OR operation. Formally, the packing function is defined as:

$$
\texttt{pack}(v, N, M) = v^\top\left[\begin{array}{c} 2 ^ {M - N} \\ 2^{M - 2N} \\ \vdots \\ 2^N \\ 1\end{array}\right] 
$$

Conversely, the unpack operation reverses the packing process and is formally defined as:

$$
\texttt{unpack}(u, N, M) = \left(u \cdot \left[\begin{array}{c} 2 ^ {N - M} \\ 2^{2N - M} \\ \vdots \\ 2^{-N} \\ 1\end{array}\right]\right) (\mod{ 2^N})
$$

Here, the $i$-th element (starting from 0) is extracted by right-shifting the packed integer $u$ by $M-N(1+i)$ bits, then applying a modulo $2^N$ to isolate the $N$-bit value. The modulo operation is equivalent to a bitwise AND with $2^N-1$, masking all but the least significant $N$ bits.

\end{document}